\documentclass[11pt]{article}

\usepackage{acl}
\usepackage{times}
\usepackage{latexsym}
\usepackage[T1]{fontenc}
\usepackage[utf8]{inputenc}
\usepackage{microtype}
\usepackage{inconsolata}
\usepackage{graphicx}
\usepackage{comment}
\usepackage{booktabs}
\usepackage{multirow}

\title{When Should We Trust Medical Image Models?\\
Evaluating Confidence Reliability under Cross-Dataset Shift}

\author{ 
Haoran Zhang \\
Boston University \\
\texttt{zhr114@bu.edu}
\And
Sarah Lam \\
Boston University \\
\texttt{sarahl@bu.edu}
\And
Yuting Lin \\
Boston University \\
\texttt{linyt@bu.edu}
\And
Nuo Chen \\
Boston University \\
\texttt{ceno@bu.edu}
\And
Mingyang Li \\
Boston University \\
\texttt{limingy@bu.edu}
  }

\begin{document}
\maketitle

\begin{abstract}
Deep learning models for chest X-ray analysis often achieve strong 
performance on benchmark datasets, but their reliability under 
cross-dataset deployment remains unclear. In clinical practice, models 
trained in one hospital system may be applied to data from another, 
where distributional and labeling differences can affect both 
prediction accuracy and model confidence.

We conduct a controlled empirical study of model confidence under 
cross-dataset transfer. We train ResNet-50 and DenseNet-121 on CheXpert 
and evaluate them on MIMIC-CXR across five experimental settings, 
focusing on Pneumothorax and Pleural Effusion. We measure predictive 
performance using AUC and confidence reliability using Expected 
Calibration Error (ECE), Brier score, and average confidence on 
incorrect predictions, all reported with 95\% bootstrap confidence 
intervals. We further evaluate post-hoc temperature scaling fit on the 
source validation set and use Grad-CAM to examine the mechanism behind 
cross-domain performance differences.

Cross-domain AUC drops 7--13 percentage points, and the best in-domain 
model is not the best cross-domain model. Source-domain temperature 
scaling produces marginal ECE changes but systematically increases 
confidence on incorrect predictions across all four model-pathology 
combinations. Grad-CAM reveals that DenseNet relies on a concentrated 
spatial prior that fails to transfer, while ResNet's more distributed 
attention generalizes better. These findings argue for cross-site 
validation and safety-specific calibration metrics in medical AI 
evaluation.
\end{abstract}

\section{Introduction}

Deep learning models have achieved strong performance in chest X-ray 
classification tasks, often reaching high AUC scores on benchmark 
datasets such as CheXpert \citep{irvin2019chexpert} and MIMIC-CXR 
\citep{johnson2019mimic}. Nevertheless, prior work has shown that 
models trained on one dataset may fail to generalize reliably to data 
from another hospital system because of distribution and labeling 
differences \citep{cohen2020limits}. In particular, cross-domain 
studies have demonstrated that concept shift and dataset-specific 
biases can significantly affect model predictions.

Although performance degradation under domain shift has been widely 
studied \citep{cohen2020limits}, less attention has been paid to how 
model confidence behaves in such settings. High accuracy does not 
necessarily imply well-calibrated confidence, and medical imaging 
models are often overconfident even when incorrect 
\citep{saha2023understanding}. In safety-critical applications, 
unreliable confidence estimates could mislead clinicians and delay 
necessary human intervention.

In this work, we focus on understanding the reliability of model 
confidence under cross-dataset transfer. We investigate whether 
confidence estimates learned on a source dataset remain trustworthy 
when evaluated on an external dataset, and whether simple post-hoc 
calibration methods such as temperature scaling \citep{guo2017calibration} 
can improve cross-domain reliability. By analyzing predictive performance 
and calibration behavior together, and by using Grad-CAM 
\citep{selvaraju2017gradcam} to examine model attention on confident 
failure cases, we characterize when medical image model outputs should 
or should not be trusted under cross-hospital deployment.

\section{Related Work}

Recent work has shown that deep learning models trained on one chest 
X-ray dataset often fail to generalize to external datasets due to 
distributional and conceptual differences \citep{cohen2020limits}. 
Concept shift arising from labeling inconsistencies and dataset-specific 
biases can significantly affect predictive performance, motivating 
evaluation of models under cross-domain deployment rather than relying 
solely on in-distribution benchmarks such as CheXpert 
\citep{irvin2019chexpert} or MIMIC-CXR \citep{johnson2019mimic}.

Beyond predictive accuracy, reliable confidence estimation is important 
in medical decision-making. Prior studies have shown that medical 
imaging models are often miscalibrated, exhibiting overconfident 
predictions even when incorrect \citep{saha2023understanding}. 
Calibration techniques such as temperature scaling have been widely 
adopted in general computer vision \citep{guo2017calibration} and have 
been studied in medical image classification, particularly under 
class-imbalance settings \citep{rajaraman2022deep}. However, most 
existing work evaluates calibration within a single dataset.

While cross-domain generalization and calibration have been studied 
separately, the interaction between domain shift and confidence 
reliability remains underexplored. In particular, it is unclear whether 
calibration learned on a source dataset remains effective when models 
are deployed on external datasets. Our work bridges this gap by jointly 
analyzing predictive performance and confidence behavior under 
cross-dataset transfer, and by using Grad-CAM \citep{selvaraju2017gradcam} 
to provide mechanism-level evidence for the cross-domain reliability gap.

\section{Method}

We designed a controlled cross-dataset evaluation setting to analyze 
model confidence under domain shift. CheXpert \citep{irvin2019chexpert} 
serves as the source dataset and MIMIC-CXR \citep{johnson2019mimic} as 
the target. To reduce task-related confounding, we focused on two 
pathologies that are consistently labeled across both datasets: 
Pneumothorax and Pleural Effusion. Restricting the study to shared 
labels controlled for task mismatch across datasets. Within the 
MIMIC-CXR evaluation set, we additionally distinguished samples 
containing only these two shared pathologies (\textit{shared\_only}) 
from samples containing other co-occurring pathologies 
(\textit{extra\_pathology}), enabling subset-level analysis of 
distribution shift.

We trained two CNN-based classifiers on the CheXpert training set: 
ResNet-50 and DenseNet-121, both initialized from ImageNet-pretrained 
weights. Including two architectures allowed us to examine whether 
confidence miscalibration under domain shift is model-specific or 
consistent across different CNN designs. Model selection and 
hyperparameter choices were performed on the CheXpert validation set. 
The trained models were then evaluated on both the CheXpert validation 
set (in-domain evaluation) and the MIMIC-CXR test set (cross-domain 
evaluation), enabling direct comparison between in-domain and 
cross-domain performance under identical model parameters.

We further applied post-hoc temperature scaling \citep{guo2017calibration} 
fit per-pathology on the CheXpert validation set. The calibrated models 
were evaluated on MIMIC-CXR without re-fitting calibration on the target 
dataset, testing whether calibration learned in the source domain 
remained effective under domain shift. To investigate the mechanism 
behind cross-domain performance differences, we applied Grad-CAM 
\citep{selvaraju2017gradcam} to high-confidence predictions on MIMIC-CXR 
across both architectures.

\section{Data}

We use CheXpert-small \citep{irvin2019chexpert} as our source dataset 
for training and in-domain evaluation. CheXpert-small is a publicly 
available subset of the CheXpert dataset released by Stanford ML Group, 
containing 224,316 chest radiographs from 65,240 patients. Each image 
is labeled for 14 radiological observations, with labels taking four 
possible values: positive (1.0), negative (0.0), uncertain ($-1.0$), 
and unmentioned (blank).

We focus on two pathologies that are consistently labeled across both 
CheXpert and our target dataset MIMIC-CXR: Pneumothorax and Pleural 
Effusion. These pathologies show contrasting prevalence in CheXpert — 
Pleural Effusion is relatively common while Pneumothorax is relatively 
rare — allowing us to examine whether calibration behavior varies with 
class frequency.

\paragraph{CheXpert preprocessing.} We applied the following steps:

\begin{itemize}
    \item Filtered to retain only frontal-view images (AP/PA), excluding 
    lateral views. This reduced the training set from 223,414 to 
    191,027 samples.
    \item Applied the U-Ones policy for uncertain labels: all $-1.0$ 
    entries were mapped to $1.0$ (positive), and all blank (unmentioned) 
    entries were mapped to $0.0$ (negative). This follows the label 
    strategy recommended in the original CheXpert study.
    \item Resized images to 256 pixels on the shorter side, then 
    center-cropped to $224 \times 224$ pixels for validation, or 
    random-cropped for training.
    \item Normalized pixel values using ImageNet channel-wise mean and 
    standard deviation, as our models are initialized from 
    ImageNet-pretrained weights.
\end{itemize}

The CheXpert-small validation set contains 202 frontal-view samples 
after filtering.

\paragraph{MIMIC-CXR for cross-domain evaluation.} For cross-domain 
evaluation, we use MIMIC-CXR \citep{johnson2019mimic} (v2.1.0, accessed 
via PhysioNet credentialed access). We use the official test split, 
applying the same preprocessing pipeline as for CheXpert. After 
filtering for frontal views and joining with CheXpert-style auto-labels 
(\texttt{mimic-cxr-2.0.0-chexpert.csv}), the test set contains 1,835 
samples. For per-pathology evaluation, we exclude samples with uncertain 
($-1.0$) labels for that pathology, yielding 1,816 valid samples for 
Pneumothorax and 1,702 for Pleural Effusion.

We additionally partition the cross-domain test set into two subsets 
based on co-occurring pathologies: \textit{shared\_only} ($N=149$), 
containing samples whose other 12 pathology labels are all 0 or blank, 
and \textit{extra\_pathology} ($N=1{,}686$), containing samples with at 
least one other confidently positive pathology label. This partition 
enables analysis of how distributional structure within the target 
domain affects cross-domain reliability.

\section{Experimental Setup}

\subsection{Model Architectures}

We trained two CNN-based classifiers under matched conditions: ResNet-50 
and DenseNet-121, both initialized from ImageNet-pretrained weights 
(\texttt{ResNet50\_Weights.IMAGENET1K\_V2} and 
\texttt{DenseNet121\_Weights.IMAGENET1K} from torchvision). For each 
model, we replaced the final classification layer with a linear layer 
outputting 2 logits, corresponding to Pneumothorax and Pleural Effusion. 
We treat the task as multi-label binary classification, where each 
pathology is predicted independently.

Including two architectures allows us to examine whether confidence 
miscalibration and cross-domain performance behavior are model-specific 
or consistent across CNN designs. Identical training conditions ensure 
that any observed differences between the two models reflect 
architectural properties rather than training-related confounds.

\subsection{Training Details}

Both models were trained on the CheXpert-small training set (191,027 
frontal-view images) using identical hyperparameters:

\begin{itemize}
    \item Loss function: BCEWithLogitsLoss (binary cross-entropy with logits)
    \item Optimizer: Adam with learning rate $1 \times 10^{-4}$
    \item Batch size: 32
    \item Training epochs: 10
    \item Data augmentation (training only): random crop, random 
    horizontal flip, random rotation ($\pm 10^{\circ}$)
\end{itemize}

Model selection was performed on the CheXpert validation set using mean 
AUC across both pathologies. The best checkpoint for each architecture 
was saved at the epoch with the highest validation mean AUC.

\subsection{Calibration and Evaluation}

We applied post-hoc temperature scaling per pathology on the CheXpert 
validation set, fitting the temperature parameter $T$ by minimizing 
binary cross-entropy via L-BFGS optimization. The fitted temperatures 
were then applied directly to MIMIC-CXR logits without re-fitting on 
the target domain, testing whether source-domain calibration transfers 
under deployment shift.

We evaluate predictive performance using Area Under the ROC Curve (AUC), 
and confidence reliability using three calibration metrics: Expected 
Calibration Error (ECE) with 15 bins, Brier score, and average 
confidence on incorrect predictions. All metrics are reported with 
95\% bootstrap confidence intervals computed from 2,000 resamples.

For mechanism analysis, we applied Grad-CAM 
\citep{selvaraju2017gradcam} to high-confidence MIMIC-CXR predictions, 
targeting the final convolutional block of each architecture 
(\texttt{layer4} for ResNet-50, \texttt{features.denseblock4} for 
DenseNet-121). We examined three case categories per model: 
high-confidence true positives ($p \geq 0.9$, label$=1$), 
high-confidence false positives ($p \geq 0.9$, label$=0$), and 
high-confidence false negatives ($p \leq 0.1$, label$=1$).

\section{Results}

We organize results around three main findings. First, cross-domain transfer 
introduces consistent AUC degradation, but the relative ordering of models 
reverses between the source and target domains (Section~\ref{sec:auc_results}). 
Second, temperature scaling, when fit on the source domain, slightly improves 
overall calibration error but systematically worsens confidence on incorrect 
predictions across all configurations (Section~\ref{sec:ts_results}). Third, 
Grad-CAM visualizations reveal that the rank reversal arises from architecture-level 
differences in attention: DenseNet relies on a concentrated spatial prior 
that fails to transfer, while ResNet learns more distributed features 
(Section~\ref{sec:gradcam_results}).

\subsection{Cross-domain AUC Degradation and Rank Reversal}
\begin{table*}[t]
\centering
\caption{Per-pathology AUC with 95\% bootstrap confidence intervals across all evaluation settings. $N$ is the number of test samples (reported as \textit{Pneumothorax\,/\,Pleural Effusion} when the two pathologies use different subsets).}
\label{tab:auc_cross_domain}
\resizebox{\textwidth}{!}{%
\begin{tabular}{llccc}
\toprule
Setting & Model & Pneumothorax AUC [95\% CI] & Pleural Effusion AUC [95\% CI] & $N$ \\
\midrule
\multirow{2}{*}{In-domain (CheXpert)} & ResNet-50 & 0.914 [0.809, 0.989] & 0.932 [0.897, 0.963] & 202 \\
 & DenseNet-121 & 0.935 [0.872, 0.989] & 0.926 [0.889, 0.959] & 202 \\
\midrule
\multirow{2}{*}{Cross-domain: all} & ResNet-50 & 0.836 [0.795, 0.875] & 0.855 [0.837, 0.873] & 1816 / 1702 \\
 & DenseNet-121 & 0.810 [0.763, 0.852] & 0.853 [0.835, 0.871] & 1816 / 1702 \\
\midrule
\multirow{2}{*}{Cross-domain: shared-only} & ResNet-50 & 0.769 [0.627, 0.888] & 0.860 [0.799, 0.916] & 147 / 139 \\
 & DenseNet-121 & 0.793 [0.675, 0.898] & 0.822 [0.749, 0.888] & 147 / 139 \\
\midrule
\multirow{2}{*}{Cross-domain: extra-pathology} & ResNet-50 & 0.839 [0.794, 0.882] & 0.855 [0.834, 0.873] & 1669 / 1563 \\
 & DenseNet-121 & 0.804 [0.751, 0.852] & 0.856 [0.837, 0.875] & 1669 / 1563 \\
\bottomrule
\end{tabular}%
}
\end{table*}

\label{sec:auc_results}
Table~\ref{tab:auc_cross_domain} reports the AUC per-pathology in all 
evaluation settings. Two consistent patterns emerge.

\textbf{AUC degrades under domain shift.} Both ResNet-50 and DenseNet-121 
show meaningful AUC drops when moving from CheXpert validation to MIMIC-CXR. 
ResNet-50 drops 7.8 percentage points on Pneumothorax (0.914 to 0.836) and 
7.7 points on Pleural Effusion (0.932 to 0.855). DenseNet-121 shows a 
similar drop on Pleural Effusion (0.926 to 0.853) but a substantially 
larger drop on Pneumothorax (0.935 to 0.810, a 12.5-point loss). High 
in-domain AUC does not directly translate to deployment performance, even 
when the source and target tasks share the same label definitions.

\textbf{The best in-domain model is not the best cross-domain model.} 
On CheXpert validation, DenseNet-121 outperforms ResNet-50 on Pneumothorax 
(0.935 vs.\ 0.914). On MIMIC-CXR, however, ResNet-50 achieves the higher 
AUC (0.836 vs.\ 0.810). This rank reversal is robust across the 
\textit{extra\_pathology} subset ($N=1{,}669$, ResNet 0.839 vs.\ DenseNet 
0.804) but does not appear on the smaller \textit{shared\_only} subset 
($N=147$, ResNet 0.769 vs.\ DenseNet 0.793). The latter contains only 20 
Pneumothorax positives, yielding wide and overlapping confidence intervals 
that limit subset-level conclusions. On Pleural Effusion, the two models 
perform within 0.003 of each other across all cross-domain settings, with 
no clear reversal. The mechanism behind the Pneumothorax-specific reversal 
is examined in Section 6.3.

\subsection{Temperature Scaling Worsens Confidence on Errors}
\begin{table*}[t]
\centering
\caption{Calibration metrics on the full cross-domain MIMIC-CXR test set, before and after temperature scaling (TS). Lower is better for ECE, Brier, and confidence on incorrect predictions. Values are point estimates with 95\% bootstrap confidence intervals.}
\label{tab:calibration_ts}
\resizebox{\textwidth}{!}{%
\begin{tabular}{lllccc}
\toprule
Model & Pathology & Calibration & ECE [95\% CI] & Brier [95\% CI] & Conf. on Incorrect [95\% CI] \\
\midrule
\multirow{4}{*}{ResNet-50} & \multirow{2}{*}{Pneumothorax} & Uncal. & 0.022 [0.018, 0.035] & 0.049 [0.042, 0.057] & 0.833 [0.807, 0.858] \\
 &  & +TS & 0.028 [0.023, 0.041] & 0.050 [0.042, 0.059] & 0.864 [0.839, 0.889] \\
\cmidrule(lr){2-6}
 & \multirow{2}{*}{Pleural Effusion} & Uncal. & 0.181 [0.163, 0.200] & 0.182 [0.171, 0.193] & 0.721 [0.708, 0.734] \\
 &  & +TS & 0.181 [0.162, 0.201] & 0.187 [0.174, 0.199] & 0.764 [0.750, 0.779] \\
\midrule
\multirow{4}{*}{DenseNet-121} & \multirow{2}{*}{Pneumothorax} & Uncal. & 0.049 [0.042, 0.060] & 0.057 [0.049, 0.065] & 0.791 [0.765, 0.818] \\
 &  & +TS & 0.046 [0.037, 0.058] & 0.057 [0.049, 0.066] & 0.819 [0.792, 0.845] \\
\cmidrule(lr){2-6}
 & \multirow{2}{*}{Pleural Effusion} & Uncal. & 0.100 [0.084, 0.119] & 0.159 [0.149, 0.169] & 0.702 [0.690, 0.715] \\
 &  & +TS & 0.105 [0.089, 0.125] & 0.162 [0.150, 0.174] & 0.741 [0.728, 0.756] \\
\bottomrule
\end{tabular}%
}
\end{table*}

\label{sec:ts_results}
We fit temperature scaling per pathology on the CheXpert validation set 
by minimizing binary cross-entropy. The optimal temperatures are all 
less than one (ResNet: $T=0.805$ for Pneumothorax, $T=0.746$ for Pleural 
Effusion; DenseNet: $T=0.820$ and $T=0.764$, respectively), indicating 
that both models are underconfident on the source validation set. We 
apply these source-domain-fitted temperatures directly to MIMIC-CXR 
logits without re-fitting, testing whether calibration learned 
in-domain transfers under deployment shift.

Table~\ref{tab:calibration_ts} reports calibration metrics before and 
after temperature scaling on the full cross-domain test set. The results 
reveal an asymmetric trade-off.

\textbf{ECE changes are small and inconsistent.} Across the four 
model-pathology combinations, ECE either marginally worsens (ResNet 
Pneumothorax: 0.022 to 0.028; DenseNet Pleural Effusion: 0.100 to 0.105) 
or remains essentially unchanged (ResNet Pleural Effusion: 0.181 to 
0.181; DenseNet Pneumothorax: 0.049 to 0.046). Bootstrap confidence 
intervals on ECE show substantial overlap between uncalibrated and 
calibrated estimates in all four cases. Source-domain-fitted temperature 
scaling does not produce consistent ECE improvement under cross-domain 
deployment.

\textbf{Confidence on incorrect predictions increases in 4 of 4 
settings.} While ECE is mixed, the safety-critical metric of confidence 
on incorrect predictions increases systematically after temperature 
scaling. ResNet Pneumothorax rises from 0.833 to 0.864; ResNet Pleural 
Effusion from 0.721 to 0.764; DenseNet Pneumothorax from 0.791 to 0.819; 
DenseNet Pleural Effusion from 0.702 to 0.741. The increase is largest 
on the two Pleural Effusion settings, where the 95\% bootstrap 
confidence intervals do not overlap. On the Pneumothorax settings, 
calibrated CIs lie largely above uncalibrated CIs but show narrow 
overlap. The direction is consistent across all four cases.

\textbf{Implication for safety-critical deployment.} Because $T<1$ pushes 
sigmoid outputs toward the extremes, temperature scaling makes confident 
predictions more confident — an effect that improves average calibration 
when the model is mostly correct, but amplifies confidence on incorrect 
predictions in proportion. In safety-critical settings such as clinical 
triage, where the cost of a confident wrong prediction is high, this 
asymmetric trade-off can outweigh the modest ECE gains reported by 
aggregate metrics. Calibration evaluation that relies on ECE alone 
risks masking this effect.

\subsection{Grad-CAM Reveals the Mechanism Behind Rank Reversal}

\begin{figure}[t]
\centering
\begin{tabular}{cc}
\includegraphics[width=0.45\columnwidth]{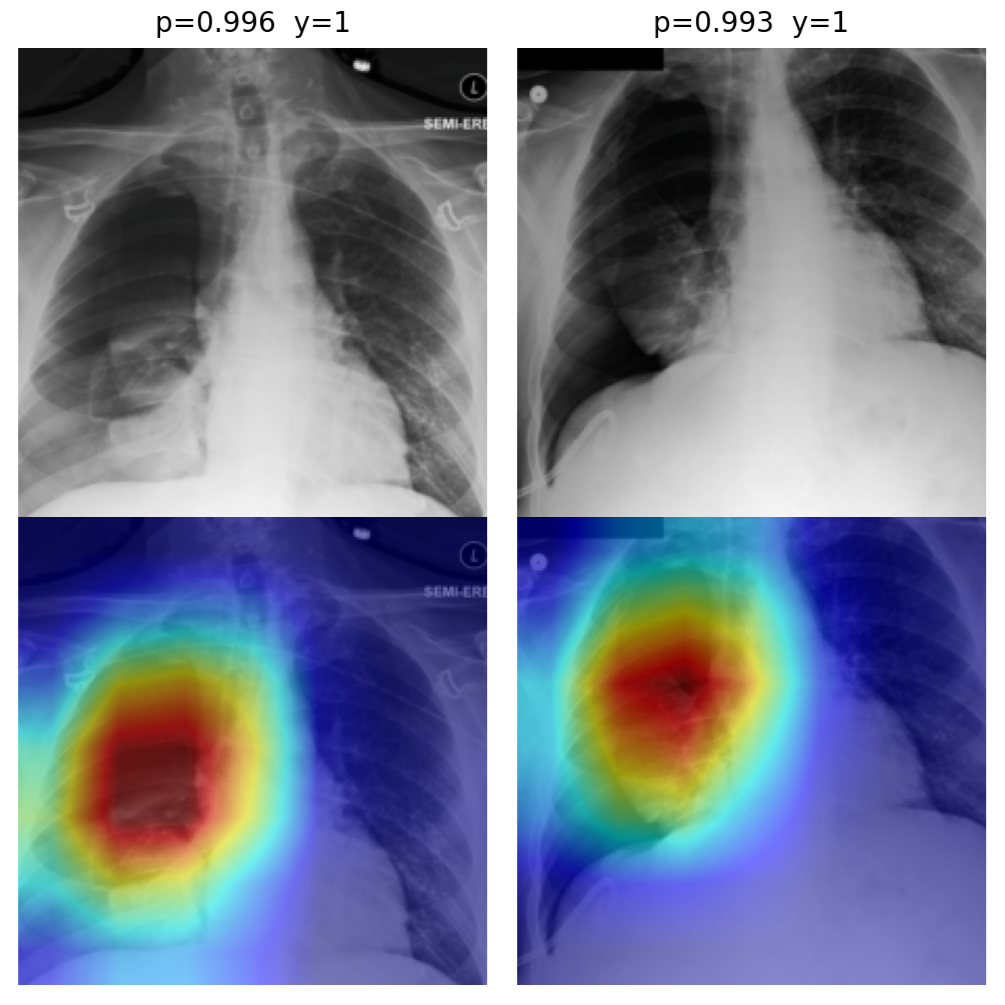} &
\includegraphics[width=0.45\columnwidth]{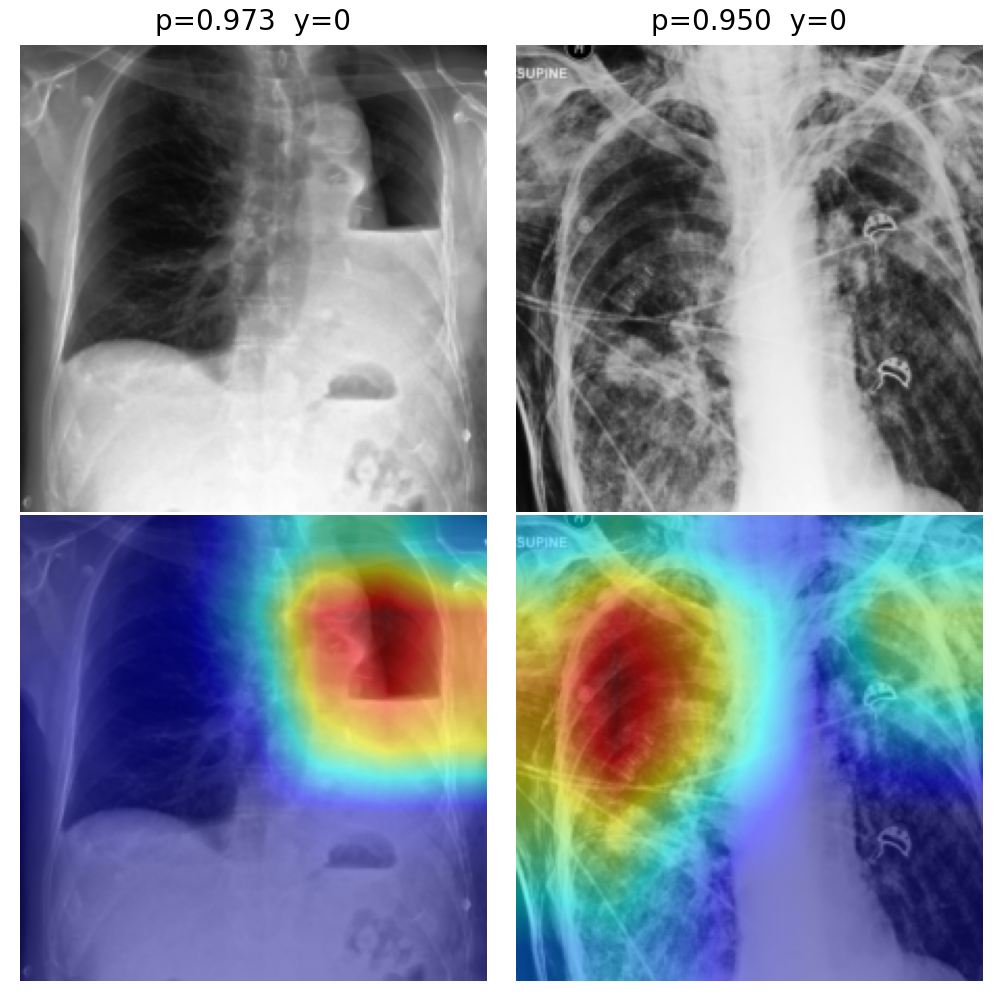} \\
\small (a) DenseNet TP & \small (b) DenseNet FP \\
\includegraphics[width=0.45\columnwidth]{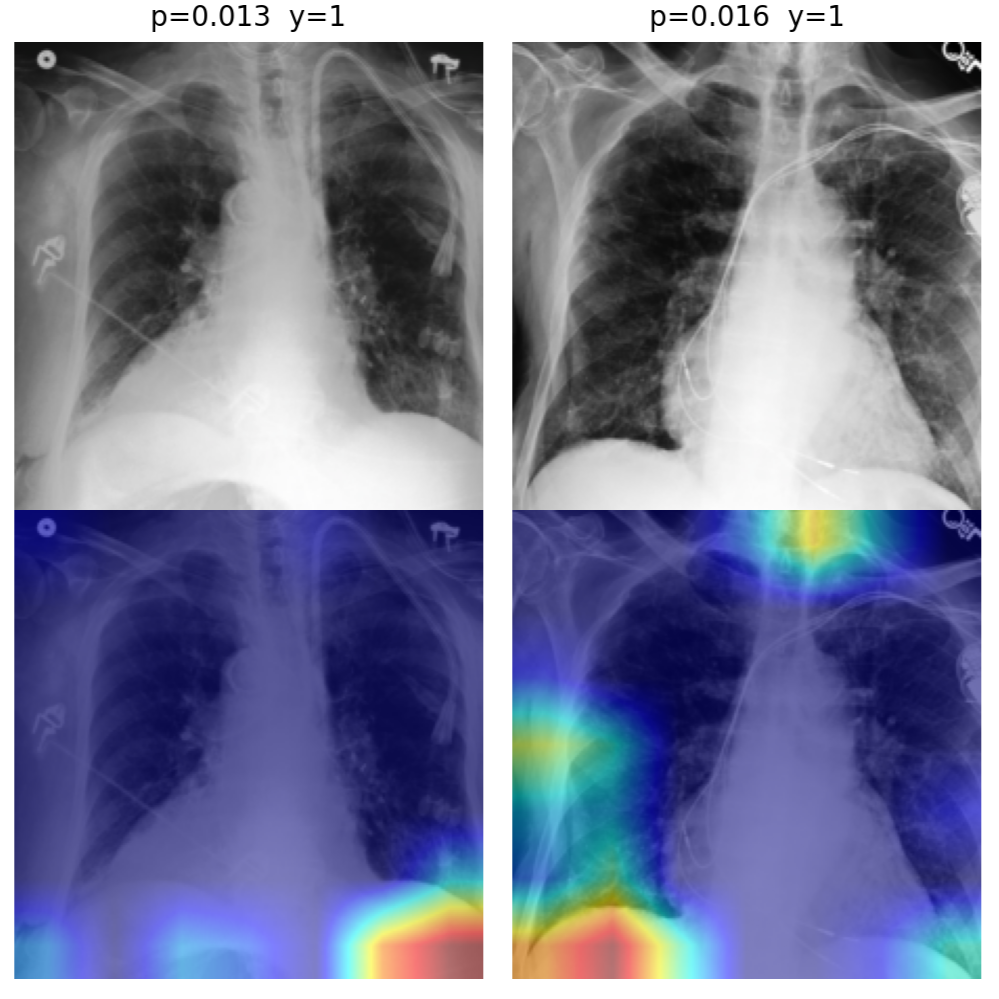} &
\includegraphics[width=0.45\columnwidth]{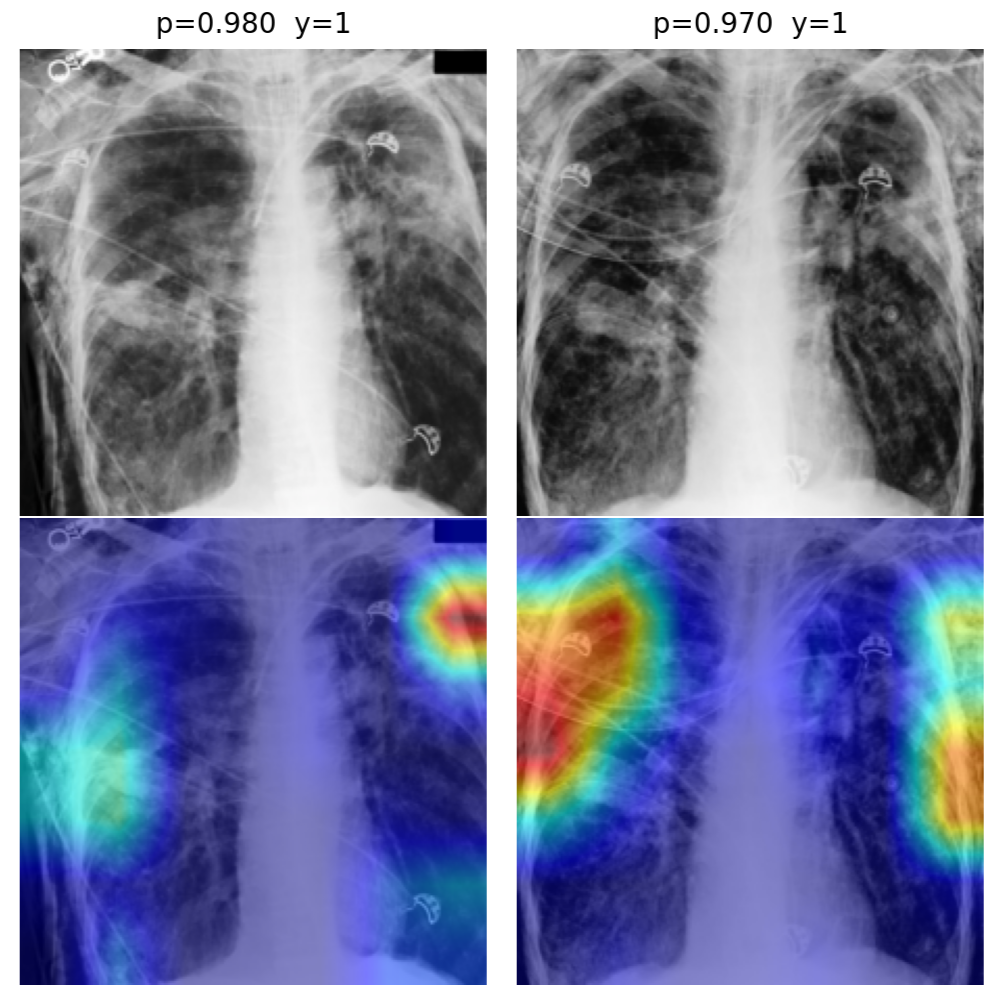} \\
\small (c) DenseNet FN & \small (d) ResNet TP \\
\end{tabular}
\caption{Grad-CAM visualizations on high-confidence MIMIC-CXR predictions. 
(a) DenseNet TPs concentrate attention on the upper right lung. 
(b) DenseNet FPs show the same pattern despite no pneumothorax present.
(c) DenseNet FNs show attention drifting to image edges, away from the lungs.
(d) ResNet TPs show more distributed attention.}
\label{fig:gradcam}
\end{figure}

\label{sec:gradcam_results}
To understand why ResNet outperforms DenseNet on cross-domain Pneumothorax 
despite the opposite ordering in-domain, we apply Grad-CAM 
\citep{selvaraju2017gradcam} to high-confidence MIMIC-CXR predictions. 
We examine three case categories per model: high-confidence true positives 
(TP, $p \geq 0.9$, label$=1$), high-confidence false positives 
(FP, $p \geq 0.9$, label$=0$), and high-confidence false negatives 
(FN, $p \leq 0.1$, label$=1$).

\textbf{DenseNet relies on a concentrated spatial prior.} 
Across DenseNet's 18 high-confidence TP cases, attention concentrates 
on the upper right lung field, with 15 of 18 cases focused on this 
region. This spatial bias is consistent with CheXpert's pneumothorax 
distribution, where iatrogenic causes (e.g., right-sided central venous 
catheter placement) are overrepresented. Critically, all 8 DenseNet FP 
cases show a visually indistinguishable attention pattern despite 
containing no pneumothorax: the model is matching a learned spatial 
template rather than detecting pathology-specific features. On confident 
FN cases, attention falls on image edges and non-anatomical regions 
rather than the lungs, indicating reliance on spurious cues to support 
the ``no pneumothorax'' decision.

\textbf{ResNet attention is more distributed.} The same analysis on 
ResNet yields a different picture: TP heatmaps distribute across both 
lung fields with varied shapes, and ResNet produces only 4 
high-confidence FP cases compared to DenseNet's 8.

\textbf{Connecting attention to rank reversal.} These observations 
suggest that DenseNet learned a feature shortcut aligned with CheXpert's 
spatial label distribution — yielding high in-domain AUC but failing 
on MIMIC-CXR, where patient population and pneumothorax etiology differ. 
ResNet, with more distributed attention, did not fit the source-specific 
template as tightly and degrades less under domain shift. Grad-CAM 
provides qualitative, suggestive evidence; quantifying attention bias 
remains future work.

\section{Discussion}

Our results suggest three main takeaways for cross-domain deployment of 
medical image classifiers.

\subsection{In-domain Benchmarks Mislead Cross-domain Reliability}

The rank reversal between ResNet-50 and DenseNet-121 on Pneumothorax 
demonstrates that in-domain AUC does not predict cross-domain robustness. 
Standard model selection workflows that pick the highest-AUC model on 
held-out validation may systematically prefer architectures that have 
overfit source-specific spatial regularities. For deployment-oriented 
medical AI evaluation, a single in-distribution test set is insufficient; 
cross-site evaluation should be standard practice, not an afterthought.

\subsection{Source-domain Calibration Does Not Transfer Safely}

Temperature scaling, fit on CheXpert validation and applied to MIMIC-CXR 
without re-fitting, provides marginal ECE improvements at best while 
systematically increasing confidence on incorrect predictions across all 
four model-pathology combinations. The mechanism is straightforward: 
$T<1$ pushes sigmoid outputs toward the extremes, which improves average 
calibration when most predictions are correct but proportionally amplifies 
confidence on errors. In safety-critical contexts such as clinical triage, 
where a confidently wrong prediction can mislead a clinician more 
dangerously than an uncertain one, this trade-off is asymmetric in the 
wrong direction.

This finding qualifies, rather than contradicts, the existing literature. 
\citet{guo2017calibration} report that temperature scaling improves ECE 
on in-distribution test data, and our in-domain results are consistent 
with theirs. Our claim is more specific: when calibration is fit on the 
source domain and applied under deployment shift, ECE-only evaluation 
masks a deterioration in safety-critical metrics. Calibration evaluation 
for medical AI should report confidence on incorrect predictions 
alongside ECE.

\subsection{Mechanism: Shortcut Features Generalize Poorly}
The Grad-CAM analysis suggests that the rank reversal we observe is not 
a random architectural artifact but a consequence of representation 
quality. DenseNet's concentrated attention on the upper right lung field 
is consistent with the iatrogenic distribution of pneumothorax cases in 
CheXpert; the same template fails on MIMIC-CXR's different patient 
population. ResNet's more distributed attention does not exploit this 
spatial regularity, and consequently transfers more reliably. This 
aligns with the broader literature on shortcut learning 
\citep{geirhos2020shortcut}, which holds that models latching onto 
easy spatial or texture features tend to underperform under 
distribution shift.

\subsection{Additional Observations}

\textbf{Pathology-specific calibration behavior.} Calibration changes 
under domain shift are not uniform across pathologies. ResNet's ECE on 
Pneumothorax actually improves cross-domain (0.080 to 0.022) — likely 
because the higher Pneumothorax prevalence in MIMIC-CXR (5.9\% vs.\ 3.5\% 
in CheXpert) compensates for the model's in-domain underconfidence. 
ResNet's Pleural Effusion ECE worsens substantially (0.100 to 0.181) in 
the same domain shift. Aggregated calibration metrics can therefore mask 
divergent per-pathology behavior.

\textbf{Subset structure and out-of-distribution intuition.} The 
\textit{shared\_only} subset, containing samples with no co-occurring 
pathologies beyond the two targets, exhibits worse ECE than the 
\textit{extra\_pathology} subset across most settings. This is 
counter-intuitive if shared-only is treated as the cleaner case. The 
explanation is distributional: CheXpert's positive cases predominantly 
contain co-occurring pathologies, so MIMIC-CXR's clean single-pathology 
samples are in fact further from the training distribution than its 
multi-morbidity samples. Notions of ``cleaner test data'' do not 
necessarily correspond to ``in-distribution.''

\subsection{Limitations and Future Work}

Our study uses a single source-target pair (CheXpert $\rightarrow$ 
MIMIC-CXR) and two pathologies; whether the rank reversal and TS 
trade-off generalize to other hospital pairs and pathologies remains 
open. Grad-CAM provides qualitative evidence; quantifying spatial bias 
(e.g., side-of-pathology distributions) and validating the iatrogenic 
etiology hypothesis would strengthen the mechanistic claims. Finally, 
we evaluate only temperature scaling; whether richer calibration methods 
such as isotonic regression or Platt scaling exhibit the same safety 
trade-off under domain shift is an open empirical question.

\section{Conclusion}

We conducted a controlled empirical study of confidence reliability 
under cross-dataset deployment in chest X-ray classification, training 
ResNet-50 and DenseNet-121 on CheXpert and evaluating on MIMIC-CXR 
across five experimental settings with bootstrap confidence intervals. 
Three findings emerge.

First, cross-domain AUC degrades by 7--13 percentage points, and the 
in-domain best model (DenseNet-121) is not the cross-domain best on 
Pneumothorax — in-distribution benchmarks are unreliable proxies for 
deployment performance. Second, source-domain temperature scaling 
provides marginal ECE changes but systematically increases confidence 
on incorrect predictions across all four model-pathology combinations, 
trading aggregate calibration improvement for safety-critical 
regression. Third, Grad-CAM analysis suggests that DenseNet's 
performance gap arises from a concentrated spatial prior that aligns 
with CheXpert's pneumothorax distribution but fails to transfer; 
ResNet's more distributed attention generalizes better.

Together, these results argue that medical AI evaluation should adopt 
cross-site validation as a default and pair calibration metrics with 
safety-specific measures such as confidence on incorrect predictions. 
Reporting in-domain AUC and ECE alone risks selecting models and 
calibration methods that look reliable on benchmarks but fail in 
deployment.

\bibliography{custom}

\end{document}